\documentclass[]{fairmeta}
\usepackage{xspace}
\usepackage{eso-pic}        
\usepackage{transparent}
\usepackage{amssymb}   
\usepackage{xcolor}

\newcommand{\ourmethod}{DirectorLLM\xspace}

\def\eg{\emph{e.g}\bmvaOneDot}

\newcommand{\bmvaOneDot}{.\xspace}

\title{%
\textcolor{blue!0!purple}{L}%
\textcolor{blue!3!purple}{l}%
\textcolor{blue!6!purple}{a}%
\textcolor{blue!9!purple}{m}%
\textcolor{blue!12!purple}{a}%
\textcolor{blue!15!purple}{\ }%
\textcolor{blue!18!purple}{L}%
\textcolor{blue!21!purple}{e}%
\textcolor{blue!24!purple}{a}%
\textcolor{blue!27!purple}{r}%
\textcolor{blue!30!purple}{n}%
\textcolor{blue!33!purple}{s}%
\textcolor{blue!36!purple}{\ }%
\textcolor{blue!39!purple}{t}%
\textcolor{blue!42!purple}{o}%
\textcolor{blue!45!purple}{\ }%
\textcolor{blue!48!purple}{D}%
\textcolor{blue!51!purple}{i}%
\textcolor{blue!54!purple}{r}%
\textcolor{blue!57!purple}{e}%
\textcolor{blue!63!purple}{c}%
\textcolor{blue!67!purple}{t}%
\textcolor{blue!70!purple}{:}%
\ DirectorLLM for Human-Centric Video Generation
}

\author[1,2,*]{Kunpeng Song}
\author[1]{Tingbo Hou}
\author[1]{Zecheng He}
\author[1]{Haoyu Ma}
\author[1]{Jialiang Wang}
\author[1]{Animesh Sinha}
\author[1]{Sam Tsai}
\author[1]{Yaqiao Luo}
\author[1]{Xiaoliang Dai}
\author[1]{Li Chen}
\author[1]{Xide Xia}
\author[1]{Peizhao Zhang}
\author[1]{Peter Vajda}
\author[2]{Ahmed Elgammal}
\author[1]{Felix~Juefei-Xu}

\affiliation[1]{GenAI at Meta}
\affiliation[2]{Rutgers University}

\contribution[*]{Work done at Meta}

\abstract{In this paper, we introduce \textbf{DirectorLLM}, a novel video generation model that employs a large language model (LLM) as the ``director'' to simulate human poses within videos. 
To enhance the authenticity of human motions in text-to-video models, we extend the LLM from a text generator to a video director and human motion simulator. 
We train the DirectorLLM to generate detailed human poses, to guide video generation,
offloading the simulation of human motion from the video generator to an LLM, effectively creating informative outlines for human-centric scenes. These signals are used as conditions by the video renderer, facilitating more realistic and prompt-following video generation. 
Experiments on automatic evaluation benchmarks and human evaluations show that our model outperforms existing ones in generating videos with higher human motion fidelity, improved prompt faithfulness, and enhanced rendered subject naturalness.}
\date{\today}

\begin{document}

\maketitle



\section{Introduction}
\label{section:intro}

\begin{figure*}
\centering
\includegraphics[width=1\linewidth]{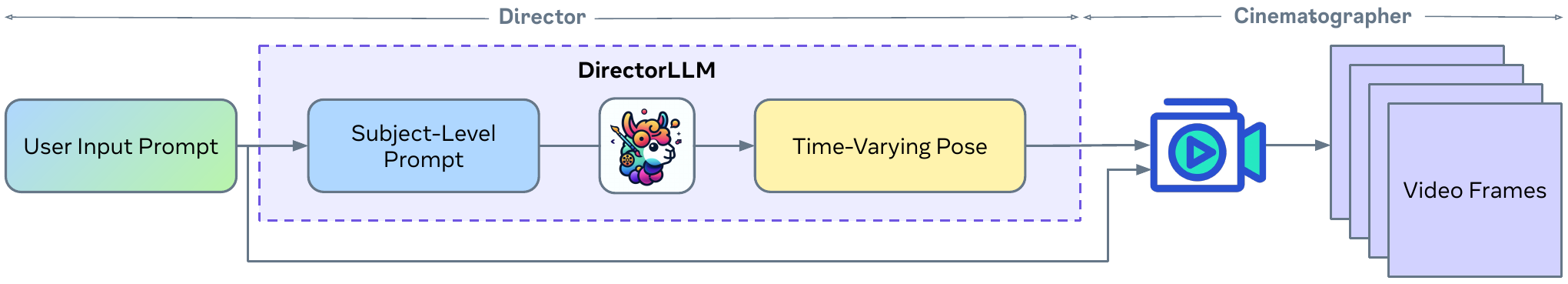}
  \caption{In the real-world movie production, the director works closely with actors to shape their portrayals, providing direction on how to deliver lines, express emotions, and move within a scene. Our vision is to \textbf{train} a large language model to assume the role of a \textbf{director} for human-centric video generation. The pipeline consists of two main parts: the \ourmethod and the video renderer (cinematographer). The \ourmethod, based on Llama 3, handles high-level scene understanding, predicting subject's location, motion, pose, and interactions within the scene based on the user input text prompt. The video renderer then generates realistic and temporally consistent video frames conditioned on the guidance provided by the \ourmethod. In this paper, we apply this concept to human-centric text-to-video generation.}
  \label{fig:teaser}
\end{figure*}

The fields of image and video generation have witnessed remarkable advancements, largely due to the emergence of diffusion-based models like DALL-E 2 \cite{ramesh2022hierarchical}, Imagen \cite{saharia2022photorealistic}, Stable Diffusion \cite{rombach2022high}, Emu~\cite{emu}, eDiff-I \cite{balaji2022ediff}, Kandinsky \cite{razzhigaev2023kandinsky}, Stable Unclip \cite{stabilityai2024}. These models empower users to create diverse and realistic visuals from text prompts.  Building on these advances, video generation models such as VideoCrafter \cite{chen2023videocrafter1} and LaVie \cite{wang2023lavie}, Sora \cite{openai_sora}, CogVideoX \cite{hong2022cogvideo}, and Movie Gen \cite{polyak2024movie} have demonstrated unprecedented capacities, achieving longer and high-quality videos across various resolutions and aspect ratios. Methods like VideoComposer \cite{wang2024videocomposer} and DreamVideo \cite{wei2024dreamvideo} integrate video generation with various control signals. Although these techniques significantly improve content generation and customization, they often face limitations in understanding human poses and motion dynamics across extended video durations.

Recent advancements in text-to-image and text-to-video generation leverage large language models (LLMs) to improve the spatial and temporal coherence of generated content. Generating complex visual dynamics directly from text prompts remains a significant challenge for diffusion models, especially when maintaining coherent spatial and temporal relationships. To address this, recent research LVD \cite{lian2023llm}, LayoutGPT \cite{feng2024layoutgpt}, attention-refocusing \cite{phung2024grounded}, and VideoDirectorGPT \cite{lin2023videodirectorgpt} have explored using LLMs like GPT-4 \cite{achiam2023gpt} and PaLM2 \cite{anil2023palm} to produce explicit spatiotemporal layouts, such as bounding boxes or frame-by-frame descriptions, derived from a single user prompt. These layouts then serve as a foundation for conditioning diffusion models, enhancing their ability to produce images and videos with coherent object relationships across frames. By transforming the narrative input into precise bounding boxes and layout structures, these methods rely on off-the-shelf LLM to simulate subject dynamics, improving motion consistency and dynamic realism.

In this work, we present \textbf{\ourmethod}, a novel T2V model designed to tackle the challenge of creating realistic and dynamic human-centric videos with improved human motions.
Existing methods \cite{chen2023videocrafter1, wang2023lavie} generally rely on diffusion models to synthesize videos directly from text prompts, requiring the model to simultaneously manage scene understanding, object interactions, and video rendering, often resulting in anatomically inaccurate human body outputs, such as broken or extra limbs, unnatural joint movements, or abrupt shifts in human motion. To address these limitations, we decouple scene understanding and pose simulation from the video generation process. Our approach consists of three main components: a specialized large language model (LLM) that interprets the text prompt to generate human poses at 1 frame per second (FPS), a compact linear diffusion model that interpolates these sparse poses to a smooth 30 FPS, enhancing temporal consistency, and a video generator (VideoCrafter \cite{chen2023videocrafter1}) enhanced with pose ControlNet \cite{zhang2023adding} to render realistic videos conditioned on both the predicted human poses and text prompt
. This structured framework allows the LLM to handle scene understanding and human motion simulation, offloading these tasks from the video generator, which can then focus on creating visually accurate frames. By integrating these specialized components, our model achieves improved realism and coherence in human-centric video generation compared to prior methods. 

Our contributions are as follows: \emph{(i)} To our knowledge, this is the first work to \textbf{train} a large language model (\eg, Llama 3 \cite{dubey2024llama}), residual VQ-VAE \cite{van2017neural}, and linear diffusion for simulating and generating human poses, specifically tailored for text-to-video generation task. \emph{(ii)} We enhance diffusion-based text-to-video (T2V) synthesis by integrating predicted human motion, leading to improved realism and coherence in human-centric videos. \emph{(iii)} Our system advances the current state of T2V synthesis, particularly in terms of human motion realism and temporal consistency.

\begin{figure*}
\centering
\includegraphics[width=1\linewidth]{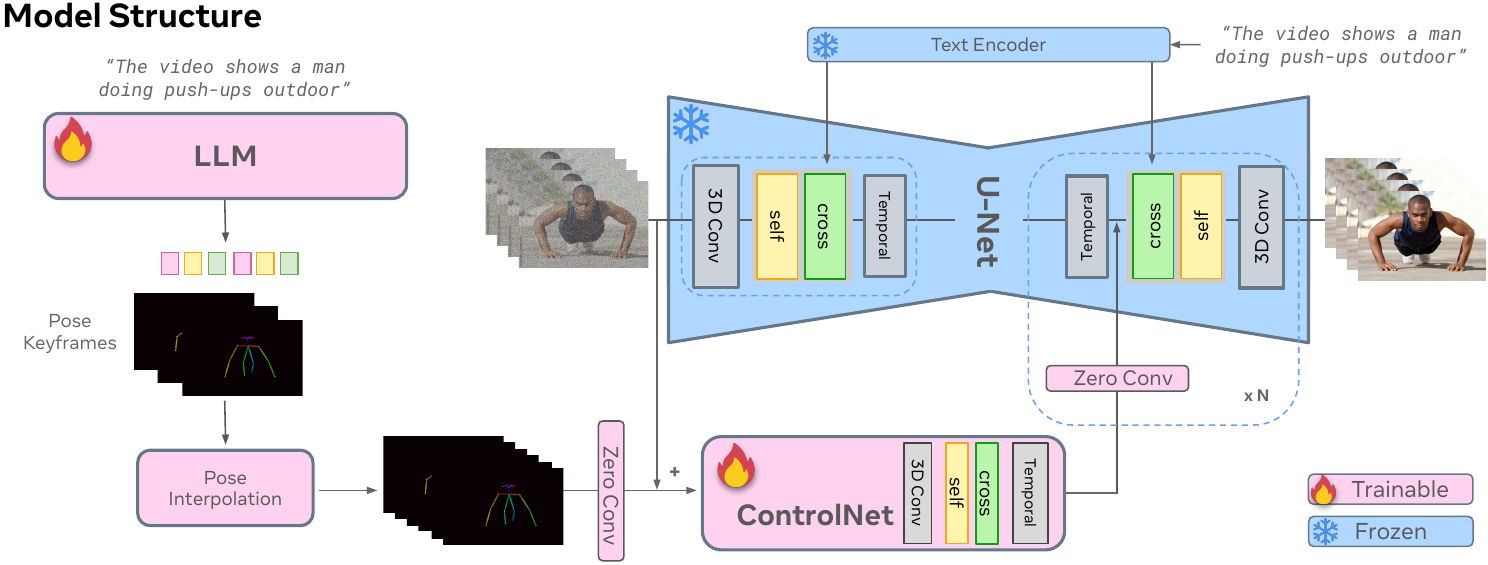}
  \caption{Overall design of our model architecture. The \ourmethod processes the text prompt to generate instance-level bounding boxes and human skeletons, effectively acting as a layout and motion planner. The linear diffusion module interpolates human skeletons, enhancing smoothness and consistency before final rendering. These outputs guide the video generation process in a U-Net-based video generator which employs ControlNet for fine-grained pose control. }
  \label{fig:pipeline}
\end{figure*}

\subsection{Related Work}
\noindent\textbf{\emph{Text-to-video (T2V) generation.}} Text-to-video (T2V) generation leverages deep learning models to interpret text input and generate video content, and it's been evolving rapidly.
Notable models, such as ModelScopeT2V \cite{wang2023modelscope} and LaVie \cite{wang2023lavie}, enhance video synthesis by integrating temporal layers within spatial frameworks to achieve more cohesive video output. VideoCrafter2 \cite{chen2023videocrafter1}, for instance, tackles the scarcity of labeled video data by making use of high-quality image datasets to improve training. Transformer-based models like Sora \cite{openai_sora} have achieved impressive results in generating videos with high visual quality, stable temporal consistency, and varied motion dynamics. Moreover, Movie Gen \cite{polyak2024movie} pushes the boundaries further, achieving exceptional quality in both video and audio generation.
%
Recent work has explored the incorporation of explicit motion controls.
Notable works include AnimateDiff \cite{guo2023animatediff}, VideoComposer \cite{wang2024videocomposer}, CameraCtrl \cite{he2024cameractrl}, Direct-A-Video \cite{yang2024direct}, and MotionCtrl \cite{wang2024motionctrl}, which design specific modules to encode camera trajectories.
Trajectory-based motion guidance is used in works such as DragNUWA \cite{yin2023dragnuwa}, MotionCtrl \cite{wang2024motionctrl}, Motion-I2V \cite{shi2024motion}, Tora \cite{zhang2024tora}, and DragAnything \cite{wu2025draganything}, while box-based motion guidance is used by TrailBlazer \cite{ma2023trailblazer} and Boximator \cite{wang2024boximator}. 
Additionally, methods like MotionDirector \cite{zhao2023motiondirector} use reference videos to extract motion patterns for generation.
In contrast to these models, which require user-provided motion signals, our model takes care of generating human poses.
, enabling fully text-driven human-centric video generation.




\noindent\textbf{\emph{LLM in image and video generation.}}
Several recent text-to-image models have incorporated large language models (LLMs) to assist image and video generation. Methods like LayoutGPT \cite{feng2024layoutgpt} employ a program-guided method for layout-oriented visual planning across various domains. Additionally, Attention Refocusing \cite{phung2024grounded} introduces innovative loss functions to realign attention maps according to specified layouts. Building upon these developments, recent efforts have extended the use of LLMs to assist text-to-video models. Free-bloom \cite{huang2024free} leverages LLMs to generate detailed frame-by-frame descriptions from a single prompt, enriching the video's narrative. LLM-grounded video generation \cite{lian2023llm} and VideoDirectorGPT \cite{lin2023videodirectorgpt} propose a training-free approach to guide diffusion models using LLM-generated frame-by-frame layouts. GPT-4-Motion \cite{lv2024gpt4motion} uses GPT-4 
\cite{achiam2023gpt} to generate scripts for Blender to produce scene components, which serve as conditions for Stable Diffusion to synthesize videos. Unlike previous work that uses pre-trained and \textcolor{blue}{frozen} language models to generate layout and bounding boxes from text prompts, our model fine-tunes a large language model, Llama 3 \cite{dubey2024llama}, to predict vector-quantized tokens to simulate human poses, enabling more accurate motion understanding tailored for video generation.

\section{Method}

\subsection{Preliminaries}
\noindent\textbf{\emph{Text-to-video.}} T2V diffusion models are designed to generate video content by adding noise in a latent space and then learning to denoise it. The process starts by encoding an input video into a latent representation, \( \mathbf{z}_0 \). Noise, represented by \( \epsilon \), is incrementally added to this latent representation over time steps \( t \), creating a noisy latent \( \mathbf{z}_t \) that corresponds to different noise levels. This progression simulates a reverse Markov chain \cite{rombach2022high} process, where the diffusion model, \( \epsilon_\theta \), is trained to predict and remove the noise added to \( \mathbf{z}_t \) as it transitions back toward the clean latent representation. The model is trained using a reconstruction loss that minimizes the difference between the actual noise \( \epsilon \) and the predicted noise \( \epsilon_\theta(\mathbf{z}_t, \mathbf{c}, t) \), where \( \mathbf{c} \) represents a conditional input, such as text or an image used for guidance. This is expressed mathematically as:
\begin{align}
\mathcal{L} = \mathbb{E}_{\mathbf{z}, \epsilon \sim \mathcal{N}(0, \mathbf{I}), t, \mathbf{c}} \left[ \left\| \epsilon - \epsilon_\theta (\mathbf{z}_t, \mathbf{c}, t) \right\|_2^2 \right]
\end{align}

\noindent\textbf{\emph{Motion generation. }} 
Prior works on diffusion have explored audio to dance \cite{tseng2023edge}, text to motion \cite{zhang2022motiondiffuse,jiang2023motiongpt}, audio to gestures \cite{alexanderson2023listen, ao2023gesturediffuclip, yu2023talking} and speech to human pose \cite{ng2024audio}. 
Audio2Photoreal \cite{ng2024audio} presents a avatar pose generation systems conditioned on audio inputs. Given speech audio, it outputs multiple possibilities of gestural motion for an individual, including face, body, and hands.
Unlike these pose models, our method focuses on generating dynamic human poses directly from text prompts at the context of text to video generation, leveraging a fine-tuned large language model for detailed and contextually accurate motion planning.

\subsection{Methodology}
\begin{figure*}
\centering
\includegraphics[width=1\linewidth]{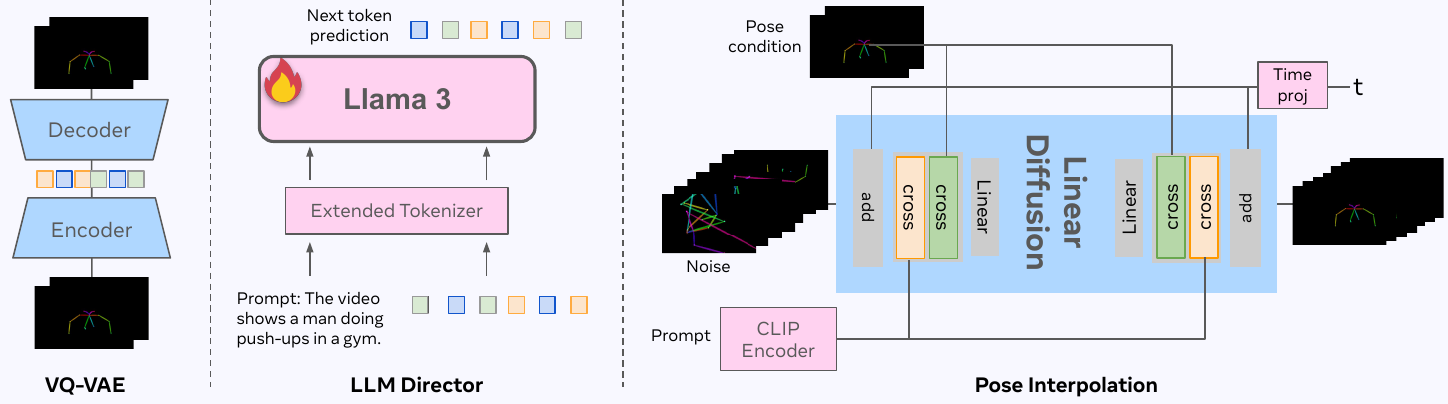}
  \caption{{Overview of \ourmethod. \textcolor{red}{Left: }The VQ-VAE module encodes human poses into discrete tokens. \textcolor{magenta}{Middle: }The Llama 3-based \ourmethod interprets text prompts to generate bounding boxes and pose tokens for each instance in the scene. \textcolor{blue}{Right: }The linear diffusion model interpolates sparse frames to dense sequences, integrating pose conditions and CLIP-encoded text prompts, producing smooth poses.}}
  \label{fig:pose}
\end{figure*}

Our model consists of three main stages: a large language model (\eg, tuned from Llama 3 \cite{dubey2024llama}) generates discretized human pose tokens from text prompts at low FPS for human motion planning, followed by a linear diffusion model for pose interpolation. The final stage uses a video generation model (VideoCrafter2 \cite{chen2023videocrafter1}) enhanced by ControlNet \cite{zhang2023adding} to produce realistic human-centric videos. 


\noindent\textbf{\emph{DirectorLLM.}}
Our \ourmethod, built on Llama 3 \cite{dubey2024llama}, plays a crucial role in scene understanding and human pose generation planning, serving as a high-level director that interprets the text prompt to produce structured, instance-level layouts for human motion. Given a prompt, the \ourmethod identifies content of the requested video in the scene and predicts corresponding human pose tokens at 1 fps. To make these outputs compatible with our generative pipeline, we convert continuous human pose vectors into discrete tokens via a residual vector quantized autoencoder (VQ-VAE \cite{van2017neural}), enabling the LLM to process human pose information as token sequences. Input sequences are \( z = (t, p) \), in which \( t = (t_1, t_2, \dots, t_n) \) are text prompt tokens, and \( p = (p_1, p_2, \dots, p_n) \) are pose tokens. The model is trained on cross-entropy loss using the standard next-token prediction framework,
\begin{align}
L = - \sum_{t=1}^{n} \log \left[p(z_{0:t-1}) = z_{t} \right]
\end{align}
where \( p(z_{0:t-1}) \) represents the predicted probability given the preceding tokens. 

Our LLM learns human pose dynamics through extensive text and video data. We show its structure in Fig.~\ref{fig:pose} \textcolor{magenta}{middle} part. By embedding human poses as tokenized predictions, our \ourmethod effectively captures the relationship between complex human dynamics and text prompts, enabling control over movement and interactions in the generated video. This component predicts 1 FPS human motions that align closely with the intended actions in the text prompt, which will be further enhance by diffusion pose interpolator.


\begin{figure*}
\centering
\includegraphics[width=1\linewidth]{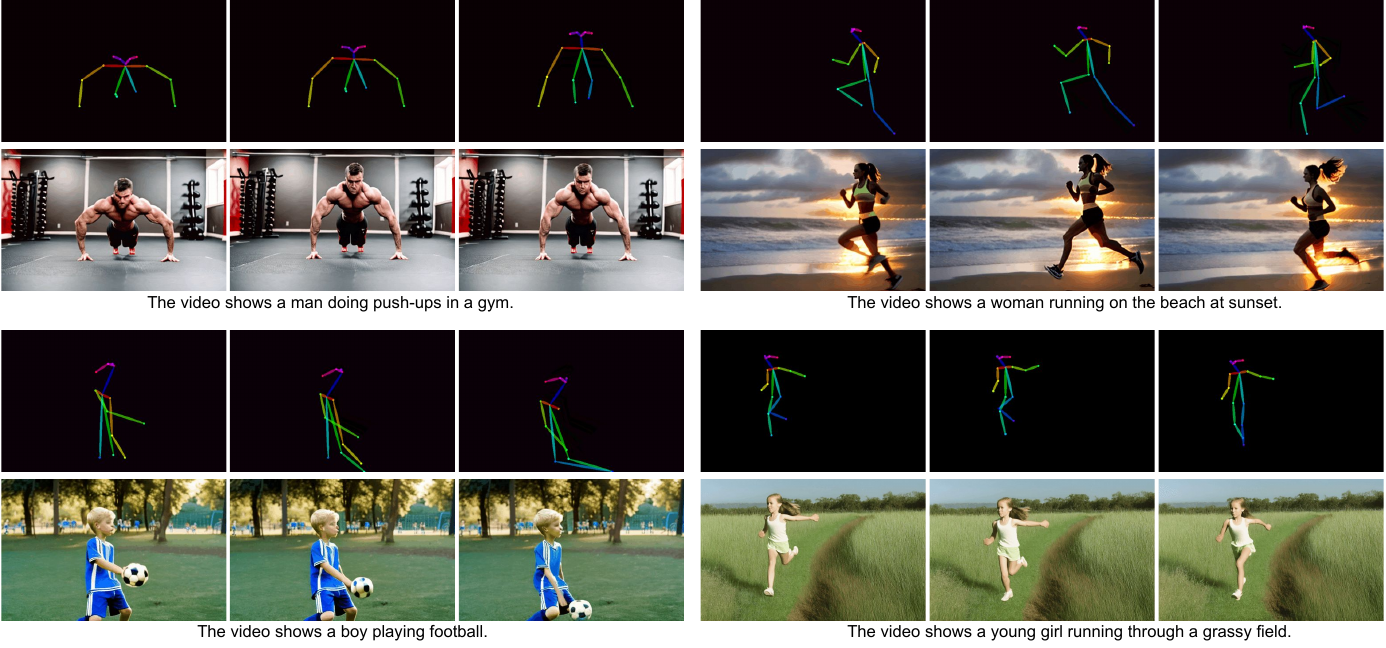}
  \caption{Examples of generated poses and videos frames from \ourmethod. Poses are generated from the LLM and interpolated by pose diffusion model. We show more pose-video sequences in the supplementary material. }
\label{fig:human_eval_vis}
\end{figure*}

\noindent\textbf{\emph{Diffusion interpolator.}}
We train a compact linear diffusion model to densify sparse poses generated by LLM into smooth and temporally consistent outputs at 30 FPS. As shown in Fig.~\ref{fig:pose} \textcolor{blue}{right} part, this model enhances the initial sparse human poses by generating intermediate frames that maintain natural motion and dynamics. We incorporate conditional pose information through cross-attention layers, allowing the model to refine pose sequences frame-by-frame based on both structural and temporal cues. To improve fidelity further, we condition the model on CLIP \cite{radford2021learning} embeddings of the text prompt, enhancing alignment between the generated dynamics and the scene description. Additionally, we employ classifier-free guidance \cite{ho2022classifier} by intermittently replacing conditions with zero tensors during training, promoting the model's ability to synthesize realistic sequences even under partial guidance. This linear diffusion process enables seamless transition from sparse layouts to dense, high-frequency poses, preparing the input for the final video generation phase.

\noindent\textbf{\emph{ControlNet video diffusion.}}
In our video generator, we build our renderer on VideoCrafter2 \cite{chen2023videocrafter1}, a text-to-video generator with a UNet architecture. We enhance following the design of ControlNet \cite{zhang2023adding} on additional human pose signals, ensuring that it adheres closely to the predicted human poses and dynamics throughout the video generation process. We incorporate zero convolution layers between ControlNet and original UNet,  which are initialized with zero weights, minimizing its interference with the original U-Net’s operations at the beginning of training. We choose a UNet structured model as our base video generator instead of a diffusion transformer (DiT) due to the substantial computational demands of DiT. Diffusion transformer requires self-attention across all spatial and temporal locations, which makes it very costly. In our design, we offload the high-level motion understanding and planning a dedicated large language model (LLM) and a pose interpolator, reducing the workload for the video generator. This division of tasks enables the UNet model to focus exclusively on rendering details based on the human poses generated by the prior components, appropriate for generating realistic, temporally consistent human-centric video frames, making it a efficient and suitable choice.

\section{Experiments}
\subsection{Dataset}
For data preparation, we start data preparation with the Shutterstock (SSTK) \cite{shutterstock2024} video dataset, generating detailed text prompts with a video captioning model to provide scene descriptions. As our task focus on human-centric video generation, we filter the dataset according to text prompt and only keep human-centric video samples. We collected videos where the same person appears across all frames following \cite{polyak2024movie}. Then we apply OpenPose \cite{cao2017realtime} on each frame to extract precise pose information. The extracted poses have 18 key points, disregrading all hand and feet points. We limits a maximum sequence length of 240 frames to capture extended scenes. During training, a subsequence of 200 frames is randomly sampled, with shorter sequences zero-padded to reach this length. From the 200 complete frames, we evenly extract 20 key frames. This thorough filtering and annotation process produces a high-quality dataset, well-suited for our model's structured scene planning and pose conditioning tasks.




\subsection{Training and Implementation Detail} 
Our training pipeline consists of four steps: (1) training a 6 depth residual vector-quantized VQ-VAE \cite{van2017neural} human poses, effectively discretizing the human poses, (2) training the large language model Director on these pose tokens, (3) training a linear diffusion model for smoothing and interpolation, and (4) training a ControlNet \cite{zhang2023adding} for the video generator.

As shown in Fig.~\ref{fig:pose} \textcolor{red}{left} part, to convert continuous human poses detected by OpenPose into discrete tokens compatible with our large language model, we train a temporally-aware vector-quantized variational autoencoder (VQ-VAE \cite{van2017neural}). Our VQ-VAE is structured as a residual linear model with a depth of six, meaning it employs six codebooks, each with a size of 512. It takes in 20 pose key frames and convert them into 120 token ids. This configuration allows for high-fidelity discretization, capturing complex pose dynamics while efficiently encoding temporal information across frames. By transforming continuous pose vectors into discretized tokens, this VQVAE enables our large language model to effectively handle and predict pose sequences as structured data within a standard text generation pipeline.

\begin{figure*}
\centering
\includegraphics[width=1\linewidth]{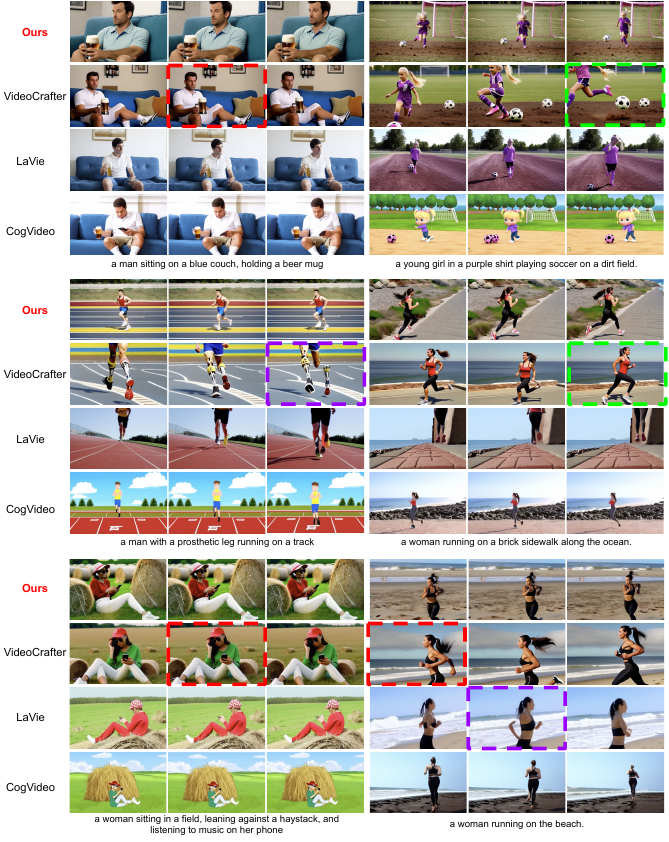}
  \caption{Qualitative comparison. Comparison of our model with baseline models (VideoCrafter, LaVie, and CogVideo) across six different video generation scenarios, each described by a text prompt. Our model consistently produces realistic and coherent human poses without visual artifacts, while the baselines frequently exhibit errors highlighted in dashed red boxes. These baseline failures include anatomical inconsistencies like extra limbs, detached body part, and abrupt direction changes.}
  \label{fig:compare}
\end{figure*}

Our large language model, fine-tuned from Llama 3~\cite{dubey2024llama}, is adapted to generate human poses tokens based on text prompts, with an extended vocabulary to accommodate newly added pose tokens. We show its structure in Fig.~\ref{fig:pose} \textcolor{magenta}{middle} part. The model is initialized from Llama 3 8B model checkpoint and all model parameters within its transformer stack are trained at this stage. The model is trained as a standard language generation model with PyTorch Lightning for 8 hours on 8 Nvidia A100 GPUs.

As shown in Fig.~\ref{fig:pose} right part, our linear interpolator integrates key frame poses and CLIP-encoded text prompts, producing smooth poses. The video ControlNet \cite{zhang2023adding} is trained on dense pose frames where we randomly subsample 16 frames and its poses for training. We initialize it by duplicating the encoder structure and parameters of the UNet in VideoCrafter, as shown in Fig.~\ref{fig:pipeline}. We keep the base UNet frozen during training and fine-tune only the pose ControlNet and zero conv layers. This approach allows our model to learn pose-specific control while maintaining the generative capabilities of the base video model. 

\subsection{Evaluation}

\noindent\textbf{\emph{Baselines and evaluation setup.}} Since we are the first work to train a LLM as director for human centric-video generation, we choose to compare with a few popular text to video generation models, including VideoCrafter \cite{chen2023videocrafter1}, LaVie-2 \cite{wang2023lavie}, and CogVideo \cite{hong2022cogvideo}. We craft a test dataset with 150 human-centric text prompts, focusing on cases with dynamic scenes and human poses.

\noindent\textbf{\emph{Qualitative evaluation.}}
In our qualitative evaluation, we show outputs from each model for six video generation scenarios, as shown in Fig.~\ref{fig:compare}. The baselines sometimes struggle with anatomical realism and temporal consistency, as indicated by several noticeable errors (highlighted in dashed colored boxes), including extra or broken limbs, and abrupt directional changes during movement. These issues reflect the baselines’ limitations in handling complex human dynamics, leading to unrealistic and visually jarring results. In contrast, our model avoids these visual artifacts, showing its improved performance in generating high-quality, human-centric videos. Due to space limits, we show more visual examples of generated pose and video frames from our model in Fig.~\ref{fig:human_eval_vis}. Further more, our \ourmethod generates 20 key pose frames which are then interpolated to 200 frames by pose interpolation model, sufficient for a video for as long as 7 seconds. We show examples of generate pose sequences and rendered RGB frames in the supplementary, demonstrating its capability of understand human dynamics and generating poses with high temporal consistency, dynamics and realism over extended durations. 

\noindent\textbf{\emph{Quantitative evaluation.}}

\begin{figure*}
\centering
\includegraphics[width=1.0\linewidth]{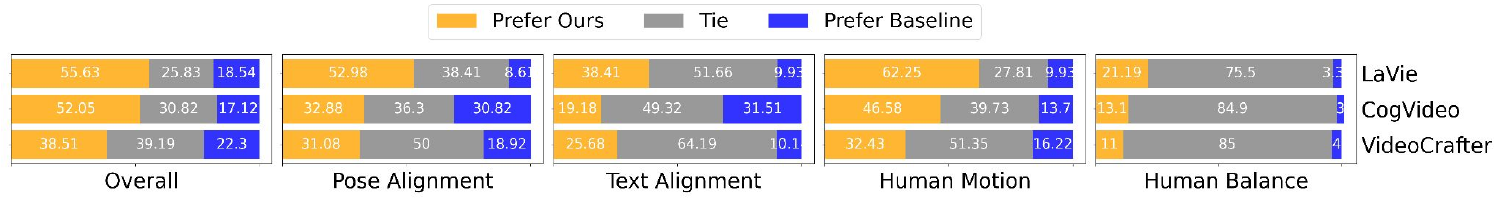}
  \caption{Human evaluations show user preference towards our method.}
  \label{fig:human_eval_bar}
\end{figure*}

We run a quantitative comparison on the entire test dataset and show the results as follows. In our experiments, we utilize VideoScore~\cite{he2024videoscore} to evaluate our text-to-video model. VideoScore is an automatic video quality assessment metric trained on VideoFeedback, a large-scale dataset containing human-provided scores for 37.6K synthesized videos from various generative models. Given its high correlation with human judgments, VideoScore serves as a reliable proxy for human evaluation, allowing us to comprehensively assess our model's performance across these crucial aspects. We show the results in Fig.~\ref{qe} for key evaluation dimensions: Visual Quality (\textbf{VC}), Temporal Consistency (\textbf{TC}), Dynamic Degree (\textbf{DD}), Text-to-Video Alignment (\textbf{CLIPvid}), and Factual Consistency (\textbf{FC}). Through quantitative evaluations, we show that our proposed method outputs realistic motion more diverse than baselines, especially in Text-to-Video Alignment (CLIPvid) score.

\begin{table}[ht]
\centering
\scalebox{1}{
    \begin{tabular}{ccccc}
    \toprule
    Metric & Ours & VideoCrafter2 & LaVie & CogVideo \\
    \midrule
    VC $\uparrow$ & \textbf{2.5205} & 2.5171 & 2.2748 & 2.2097 \\
    TC $\uparrow$ & \textbf{2.4578} & 2.4452 & 2.1527 & 2.0874 \\
    DD $\uparrow$ & \textbf{3.2663} & 3.2294 & 3.0343 & 3.0551 \\
    CLIPvid $\uparrow$ & \textbf{3.2471} & 3.1998 & 2.8947 & 2.9583 \\
    FC $\uparrow$ & \textbf{2.3598} & 2.3238 & 2.1117 & 2.1591 \\
    \midrule
    average & \textbf{2.7703} & 2.7430 & 2.4936 & 2.4939 \\
    \bottomrule
    \end{tabular}
}
\caption{Comparison of different models across various metrics. Evaluated on human-centric dynamic prompts.}
\label{qe}
\end{table}


\noindent\textbf{\emph{Human evaluation.}}
Since there is no automatic video metric on the quality of human pose dynamic, we apply human evaluation specifically for this purpose. We compare the best results from each method and make a questionnaire for human evaluators. Specifically, we run blind A/B comparisons between videos generated from each model and ask questions on: (1) overall quality (\textbf{Overall}), (2) prompt alignment for the human pose (\textbf{Pose Alignment}), (3) prompt alignment for the video (\textbf{Text Alignment}), (4) generation quality for the dynamic human motion (\textbf{Human Motion}), and (5) generation quality for static human balance (\textbf{Human Balance}). For {Text Alignment}, annotators are instructed to consider all prompt specifications and the better one would be one that achieves all specifications in the prompt. {Human Motion} evaluates the realism of dynamic human movement within the generated video, focusing on natural and coherent motion patterns. {Human Balance} assesses the static realism of human figures, ensuring accurate proportions and anatomical balance, such as correct head-to-body ratios and plausible body alignment. For each pair of comparisons, we would randomize A/B and use 3 raters per comparison, and use the majority vote as the final decision. The human evaluation setup was inspired by the video generation model \cite{polyak2024movie}.

Fig.~\ref{fig:human_eval_bar} shows the results of human evaluation comparing our model to baselines (LaVie \cite{wang2023lavie}, CogVideo \cite{hong2022cogvideo}, and VideoCrafter \cite{chen2023videocrafter1}) across five categories. Our model consistently outperformed baselines in key metrics, particularly in Pose Alignment and Human Motion, demonstrating its superior capability in generating human-centric and text-aligned video content. This aligns with our core motivation to improve the realism and accuracy of human representations in generated videos by offloading high-level scene understanding and pose prediction to a dedicated language model. This enhances both the static and dynamic realism of human figure, fulfilling our objective to produce better human-centric videos.

\subsection{Ablation and Analysis}
To better understand the contributions of each component in our model, we conducted ablation studies on three key aspects: vector quantization (VQ-VAE), the use of basic interpolation versus our linear diffusion model, and the impact of text conditioning in the linear diffusion model.

\noindent\textbf{\emph{Residual vector quantization. }}
In this experiment, we ablate our design choice of residual pose VQ-VAE. The goal is to evaluate whether our 6 depth residual autoencoding benefits LLM’s predictive performance and video quality. As shown in Fig.~\ref{qe_ab}, without residual autoencoding, the VQ-VAE couldn't accurately reconstruct human pose, significantly degrading pose prediction accuracy and introducing inconsistencies in generated videos. Residual structure in our VQ-VAE, with a depth of 6, plays a crucial role in preserving pose fidelity, directly enhancing both LLM prediction accuracy and overall video quality.

\begin{table}[ht]
\centering
\scalebox{1}{
    \begin{tabular}{cccccc}
    \toprule
      & VC $\uparrow$ & TC $\uparrow$ & DD $\uparrow$ & CLIPvid $\uparrow$ & FC $\uparrow$\\
    \midrule
    w/o Residual VQ & 2.380 & 1.815 & 3.007 & 2.773 & 1.874\\
    w/o Pose Diffusion & 2.393 & 2.197 & 3.268 & 2.861 & 2.025\\
    w/o Text Condition  & 2.445 & 2.382 & 3.255 & 3.203 & 2.331\\
     Ours & \textbf{2.520} & \textbf{2.457} & \textbf{3.266} & \textbf{3.247} & \textbf{2.359} \\
    \bottomrule
    \end{tabular}
}
\caption{Comparison of different models across various metrics. Our ablation demonstrates the importance of each component. Residual vector quantization improves pose reconstruction and LLM predictions, the linear diffusion model ensures temporal consistency and smooth motion, and text conditioning enhances alignment with prompts. }
\label{qe_ab}
\end{table}

\noindent\textbf{\emph{Linear diffusion model. }}
To assess the impact of our diffusion-based pose interpolation model, we replaced it with direct pose interpolation by linearly interpolating the 20 key frames to generate all 200 frames. These frames are then rendered to videos with our video generator. As shown in Table.~\ref{qe_ab}, the direct interpolation lacks authentic human dynamics and affects scores such as dynamic degree. Linear diffusion model outperformed basic interpolation in maintaining temporal consistency and generating smooth motion transitions, particularly in scenes with dynamic human movements.

\noindent\textbf{\emph{Text condition in linear diffusion model. }}
We also tested the effect of adding text-conditioned CLIP embeddings in the linear diffusion model by training it with and without the text conditions. As shown in Table.~\ref{qe_ab}, without text conditioning, the generated videos exhibited reduced alignment with the original text prompts, leading to inconsistencies in scene layout and subject interaction. 

\section{Discussions}
As far as our best knowledge goes, we are the first work that have showcased the effectiveness of offloading human motions simulation to an LLM for human centric T2V generation. Our approach addresses the broader challenge of human video generation with dedicated motion reasoning. We believe this opens a way to motion-simulated video generators, where the motion and interaction of the main characters are driven by dedicated AIs.

\section{Conclusions}
\label{sec:conclu}


We introduce \ourmethod, a text-conditioned video generation model that excels in creating temporally consistent, realistic videos, especially in complex scenes involving human interactions and detailed poses. By delegating scene understanding tasks to a specialized large language model that predicts bounding boxes and poses, our approach allows the video generator to focus solely on rendering. Our three-part design—consisting of the LLM, a linear diffusion model for smooth interpolation, and an enhanced VideoCrafter generator—enables precise, structured video synthesis that outperforms existing methods, particularly in scenarios requiring complex subject dynamics. This model highlights the benefits of separating scene comprehension from video generation, advancing the potential for controlled, high-quality human-centric video generation.

\clearpage

\bibliographystyle{assets/plainnat}
\bibliography{paper}

\newpage
\beginappendix


\section*{A. Implementation Details}

\subsection*{A1. Pre-Processing Details} 
Our video caption model generates highly structured prompts. Below is an example of full text prompt generated by video captioning model:

\noindent\textit{\textcolor{purple}{The video shows a woman running through a grassy field.} \textcolor{teal}{The woman has fair skin and long brown hair pulled back into a ponytail. She is wearing a yellow tank top and black shorts. She is running towards the right side of the frame. The background is a bright sky and a grassy hill. The woman is running through a grassy field. She is wearing a watch on her left wrist. The camera is static.}}

The \textbf{\textcolor{purple}{first sentence}}, known as the \textbf{\textcolor{purple}{"Subject-Level Prompt"}}, provides an overall video description with key information on human motions and poses. We use it as input to DirectorLLM transformer and Pose Linear Interpolator. We use full prompt as input to ControlNet Diffusion generator as it offers detailed descriptions of appearance and scene elements. We use OpenPose to extract per-frame human-poses for 25K videos. For videos with more than one human pose detected, we run a light weight human detection model InsightFace \cite{insightface} to keep the one with the largest area and ignore all others. 

\subsection*{A2. Model and Training Details} 
\noindent\textbf{Pose representation.} Keypoints for face, hand and feet are ignored, resulting in a 2D represenration with 18 keypoints per human pose. The coordinates of keypoints are flattened to 36 dimentional vectors, on which our Pose VQ-VAE and interpolation model work. Missing or un-detected keypoints are replaced with 0 values and ignored in loss function. We only keep 200 frames for each video, with shorter ones padded by zero.

\noindent\textbf{Residual VQ-VAE.}
Our residual VQ-VAE is designed to capture fine-grained details by employing a cascade of 6 codebooks of 512 tokens, progressively refining approximations for each pose. Both the encoder and decoder consist of a series of 1D convolutions with a kernel size of 2, providing a total receptive field of 8. Each pose is represented by a sequence of 6 VQ tokens, enabling detailed and structured encoding. For each video, we evenly sample 20 key frames from a total of 200 frames of human poses. These key frames can be converted to \textit{6 x 20 = 120}  tokens by VQ-VAE encoder. The model is compact, requiring about 2 hours of training on a single A100 GPU.
 
\noindent\textbf{DirectorLLM.} We leverage the advanced pre-existing knowledge from pretrained large languare model to aid in model convergence. Specifically, we initialize our DirectorLLM with Llama 3 -8B checkpoint\cite{dubey2024llama} from HuggingFace. We extend its tokenizer and embeddings to an additonal \textit{6 x 512} tokens, each of which representing a pose token from VQ-VAE codebooks. We attach \textit{6 x 20 = 120} pose tokens from VQ-VAE to the "subject-level prompt", forming a sequence as long as 300 tokens. Training code is built with Pytorch-Lightining and the model is trained as a standard language generation model for 8 hours on 8 Nvidia A100 GPUs.

\noindent\textbf{Pose Interpolation.} To densify human motion from 1 fps to 30 fps, we construct an conditional diffusion model. It is built on top of the Audio2PhotoReal \cite{ng2024audio}. We follow the standard DDPM definition of diffusion, where we add noise to pose vectors during forward pass and train the network to remove it during as the denoising pass. It takes in noise and predict 200 pose vectors, each for one frame, conditoned on 20 key frame poses. To incorporate the text prompt information, we use CLIP text encoder and apply a cross attention layer. Timestep information is incorporated with a feature-wise linear modulation (FiLM) layer, similar to Audio2PhotoReal \cite{ng2024audio}. All parameters are optimized. Training is finished in 8 hours on one single A100 GPU.  

\noindent\textbf{Text-to-video ControlNet.} We build our video renderer model on top of VideoCrafter2 \cite{chen2023videocrafter1}. A pose ControlNet \cite{zhang2023adding} module is initialized by copying its UNet encoder structure and parameters. We convert poses for 16 frames from 32 dimensional vector to image, and feed it to ControlNet. Connection layers are initialzed with zero convolution for better converging. We lock the parameters of base UNet and optimize only our pose ControlNet. We apply large gradient accumulation steps (200) and clip grad norm for smoother training. Training is finished with AdamW optimizer and a learning rate of \( 1^{-5} \) on 32 A100 GPUs in 48 hours.

\subsection*{A3. Evaluation} 
\noindent\textbf{Evaluation prompts.} To ensure a meaningful comparison with baseline models, we evaluate our approach using 150 human-centric text prompts randomly sampled from the test dataset. These prompts are carefully filtered to focus on scenarios involving complex scenes and dynamic human poses, selecting only those containing \textbf{keywords} indicative of large motions, including "walking", "hiking", "running," "yoga," "jogging," "workout," "dance," "jump", and "cycling". This filtering process is essential as it emphasizes the model's ability to handle challenging cases where realistic motion dynamics and pose accuracy are critical. We notice minimal performance gain with stationary prompts. By targeting prompts with significant motion, we highlight the strengths of our model in generating human-centric videos that require high fidelity in pose transitions and temporal consistency, areas where many baseline methods often struggle. This approach ensures that the evaluation reflects the practical utility and robustness of our model in human-centroc, motion-intensive scenarios.

\begin{figure*}
\centering
\includegraphics[width=1\linewidth]{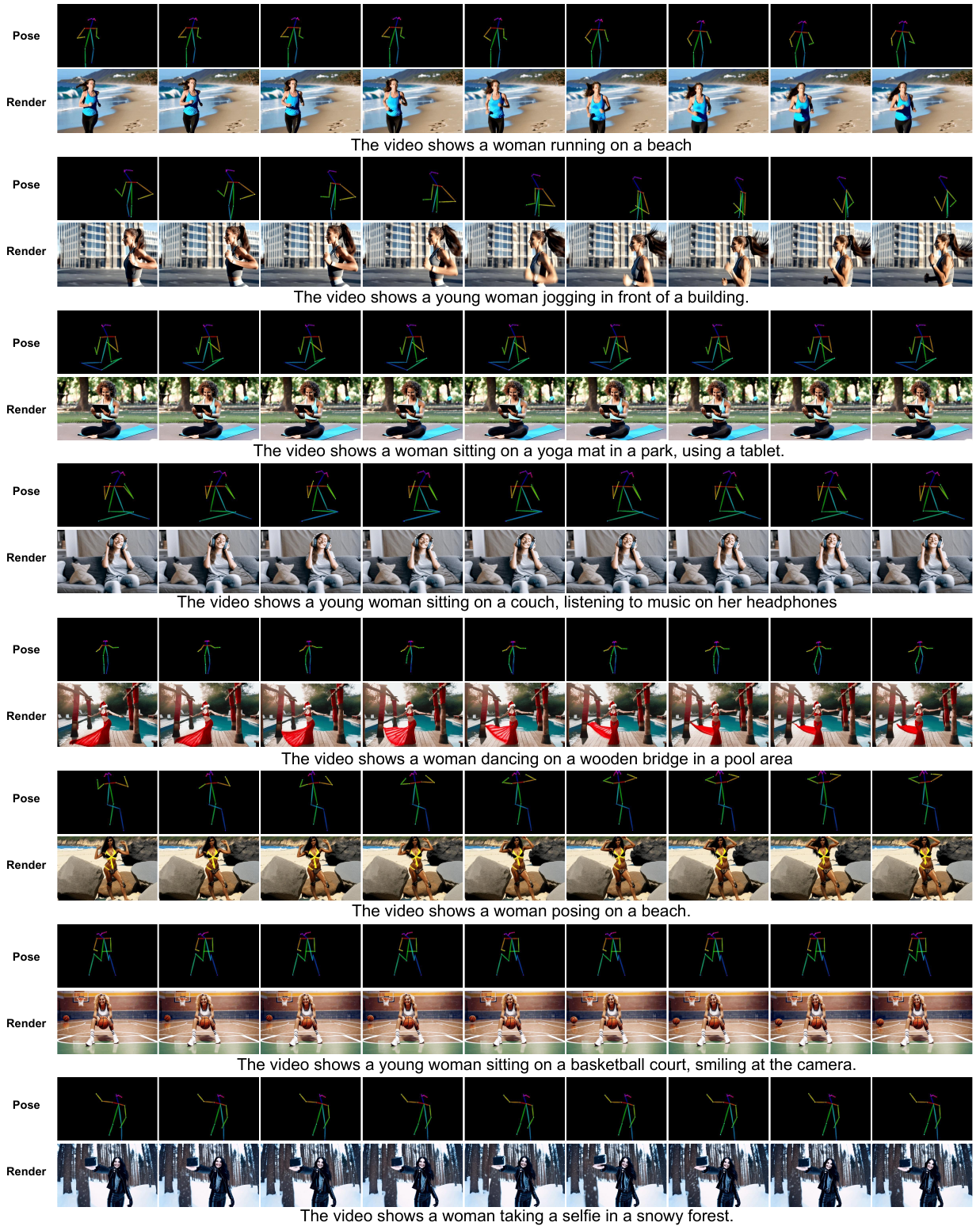}
  \caption{Generated poses and video frames.}
\label{fig:render}
\end{figure*}

\noindent\textbf{Generation pipeline.}
Generation process begins with a text prompt, which is passed to \ourmethod, to generate \textit{6 x 20 = 120} human pose tokens, represending key frame poses at 1 frame per second (fps). Pose tokens are then decoded to pose vectors with the decoder of VQ-VAE. The sparse pose vectors are then processed by a linear diffusion model. This model interpolates the 20 key frame sparse poses into 200 frames dense sequences, ensuring temporal consistency and smooth motion transitions. The linear diffusion model is also conditioned on text embeddings from CLIP to align motion dynamics with the prompt. The video generator, based on VideoCrafter and ControlNet, takes dense pose along with the full text prompt as input, producing high-quality and anatomically accurate video frames. However, due to limitations, only 16 frames are generated per inference time.

We use FIFO \cite{kim2024fifo} to bridge the gap between short-frame limitations and long-sequence requirements, which will be covered in section \textcolor{red}{C}.

\section*{B. More Results from Our Model}
We present additional visual results to demonstrate the effectiveness of our model in this section. \cref{fig:pose} upper panel showcases the sequence of human poses predicted by the LLM Director, highlighting its ability to generate accurate and temporally coherent pose layouts based on text prompts. \cref{fig:render} pairs these pose layouts with their corresponding video renderings, illustrating how the poses guide the video generator to produce realistic, high-quality human motions and interactions. The alignment between the predicted poses and the rendered visuals demonstrates the strength of our multi-stage pipeline


\begin{figure*}
\centering
\includegraphics[width=1\linewidth]{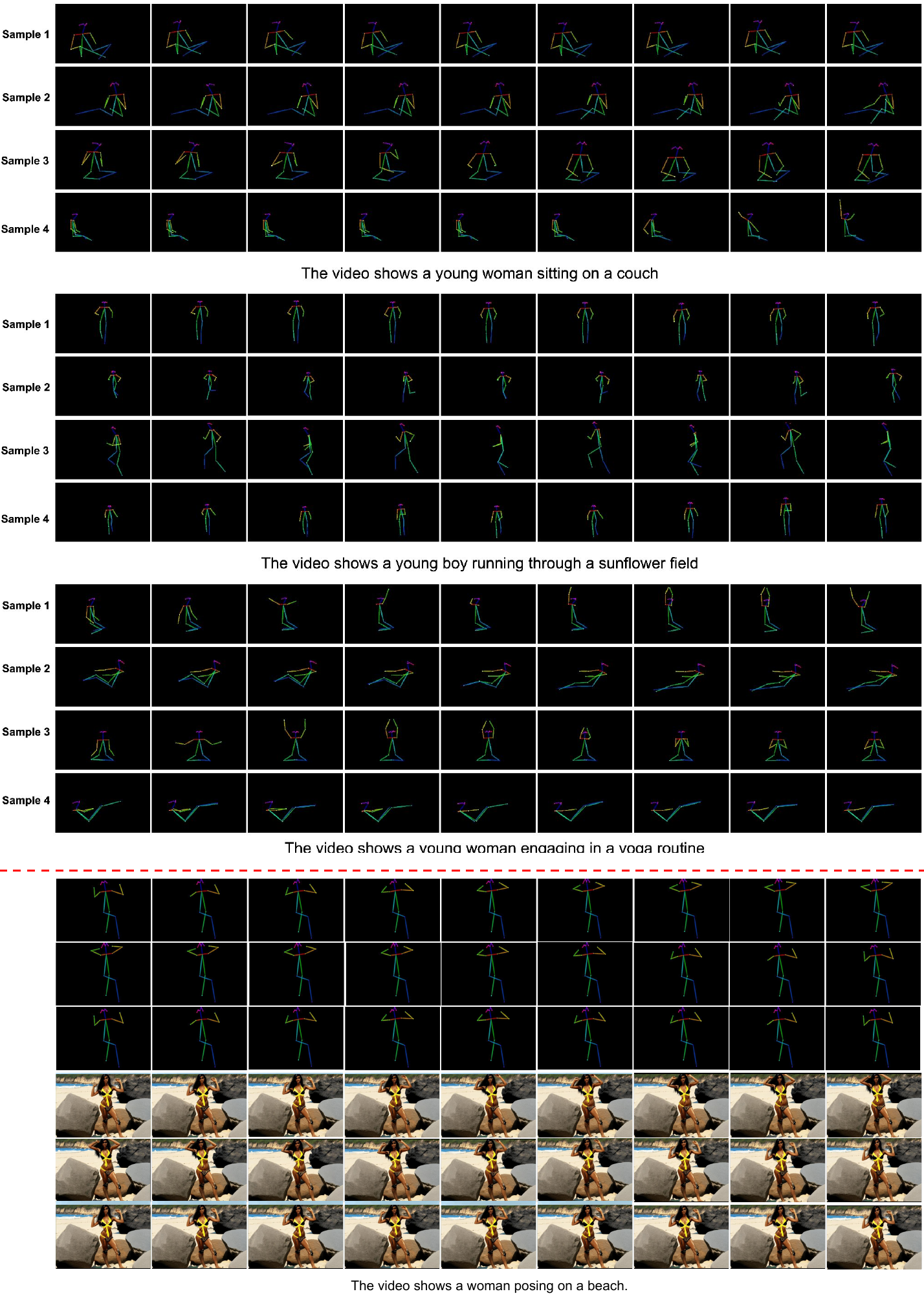}
  \caption{Generated samples from \ourmethod after pose interpolation. Each prompt showcases four representative samples to highlight the model can generate various poses given the same prompt.}
  \label{fig:pose}
\end{figure*}


\section*{C. Extending Video Length}
Our pose interpolation model generates 200 pose frames, while VideoCrafter with ControlNet is limited to generating 16 frames at a time. To extend the length of the generated video and render all 200 frames, we apply the FIFO-Diffusion \cite{kim2024fifo} approach. This method enables infinite-length video generation by employing diagonal denoising, which processes consecutive frames in a queue. During this process, fully denoised frames at the head of the queue are dequeued, while new noise frames are enqueued at the tail for processing. To mitigate the training-inference gap caused by this iterative strategy, FIFO incorporates latent partitioning and lookahead denoising techniques \cite{kim2024fifo}, ensuring consistency across extended sequences. FIFO uses the exact same model checkpoint as our base model. Simply adding ControlNet modules to the FIFO code allows us to generate long videos with pose condition awareness, leveraging VideoCrafter’s capabilities to produce human-centric videos for as long as 200 frames. A sample is attached in \cref{fig:pose} lower panel. 


\section*{D. About User Study}

We design a complex and complete evaluation pipeline with 5 questions. Specifically, we run blind A/B comparisons between videos generated from each model and ask questions on: (1) overall quality (\textbf{Overall}), (2) prompt alignment for the human pose (\textbf{Pose Alignment}), (3) prompt alignment for the video (\textbf{Text Alignment}), (4) generation quality for the dynamic human motion (\textbf{Human Motion}), and (5) generation quality for static human balance (\textbf{Human Balance}). To reduce potential biases, we recruit annotators from various countries with diverse backgrounds. Each annotator is presented with randomly paired video samples without any knowledge of the generating model to maintain impartiality. In total, we collect over 5K ratings, providing a robust and comprehensive assessment of the comparative performance of the models.

\end{document}